\documentclass[conference]{IEEEtran}
\IEEEoverridecommandlockouts
\usepackage{cite}
\usepackage{amsmath,amssymb,amsfonts}
\usepackage{algorithmic}
\usepackage{graphicx}
\usepackage{textcomp}
\usepackage{xcolor}

\usepackage{wrapfig}

\usepackage{url}

\usepackage{listings}	
\definecolor{dkgreen}{rgb}{0,0.6,0}
\definecolor{gray}{rgb}{0.5,0.5,0.5}
\definecolor{mauve}{rgb}{0.58,0,0.82}
\usepackage{pgfplots}
\usepackage{subcaption}
\usepackage{float}

\def\BibTeX{{\rm B\kern-.05em{\sc i\kern-.025em b}\kern-.08em
    T\kern-.1667em\lower.7ex\hbox{E}\kern-.125emX}}

\begin{document}

\title{\texttt{jsdp}: a Java Stochastic DP Library
}

\author{\IEEEauthorblockN{Roberto Rossi}
\IEEEauthorblockA{\textit{University of Edinburgh Business School} \\
Edinburgh, UK \\
roberto.rossi@ed.ac.uk}
}

\maketitle

\begin{abstract}
Stochastic Programming is a framework for modelling and solving problems of decision making under uncertainty. 
Stochastic Dynamic Programming is a branch of Stochastic Programming that takes a ``functional equation'' approach to the discovery of optimal policies.
By leveraging constructs --- lambda expressions, functional interfaces, collections and aggregate operators --- implemented in Java to operationalise the MapReduce framework, \texttt{jsdp} provides a general purpose library for modelling and solving Stochastic Dynamic Programs. 
\end{abstract}

\section{Introduction}

Stochastic Programming is a framework for modeling and solving problems of decision making under uncertainty. Stochastic Dynamic Programming, originally introduced by Bellman in his seminal book ``Dynamic Programming'' \cite{Bellman:1957}, is a branch of Stochastic Programming that deals with multistage decision processes and takes a ``functional equation'' approach to the discovery of optimal policies.

In ``Reminiscences about the origins of linear programming'' \cite{Dantzig_1982} Dantzig stressed the pivotal role that declarative modelling --- the ability to state objectives and constraints in clear terms, as opposed to developing ad-hoc heuristics for generating solutions - played in the development of (Linear) ``Programming;'' a tool he developed to ``compute more rapidly a time-staged deployment, training and logistical supply program.''

Powell recently remarked that the field of mathematical programming has benefited tremendously from a common canonical framework consisting of decision variables, constraints, and an objective function, while stochastic optimisation has not enjoyed this common declarative framework \cite{Powell_2016}. In other words, whilst declarative modelling has been a staple of Operations Research since these early days, to the best of our knowledge, there is limited availability of declarative modelling frameworks for Stochastic Dynamic Programming. 

By leveraging declarative constructs --- lambda expressions, functional interfaces, collections and aggregate operators --- that have been implemented in Java to operationalise the MapReduce \cite{10.1145/1327452.1327492} framework, \texttt{jsdp} aims to tackle this gap and provide a general purpose Java library\footnote{The \texttt{jsdp} library is available at: \url{http://gwr3n.github.io/jsdp/}.} for modelling and solving Stochastic Dynamic Programs. 

Our library exploits built-in parallelisation available in Java and paves the way towards a high level modelling framework that may seamlessly leverage BigData computing environments such as Hadoop and Apache Spark. In this sense, it contributes to Freuder's quest in Pursuit of the Holy Grail \cite{10.1145/242224.242304}: ``The user states the problem, the computer solves it.''

The rest of this work is structured as follows. In Section \ref{sec:sdp} we outline the general structure of a stochastic dynamic program; in Section \ref{sec:language_constructs} we survey relevant language constructs leveraged by the library; in Section \ref{sec:sdp_java} we illustrate how these language constructs can be used to implement a general purpose approach to modelling and solving stochastic dynamic programs; in Section \ref{sec:jsdp} we introduce our new library; in Section \ref{sec:applications} we briefly survey a range of applications; finally, in Section \ref{sec:related_works} we survey related works and draw conclusions.

\section{Structure of a stochastic dynamic program}\label{sec:sdp}

Stochastic dynamic programming \cite{Bellman:1957} is a branch of stochastic programming that deals with multistage decision processes and takes a ``functional equation'' approach to the discovery of optimal policies. In what follows, we adopt the canonical framework in \cite{Powell_2016} to capture the dimensions of a stochastic optimisation problem. 
Without loss of generality, we will consider a cost minimisation setting. To model a problem via stochastic dynamic programming one has to specify:
\begin{itemize}
\item a {\bf planning horizon} comprising $n$ periods;\index{planning horizon}
\item the finite set $S_t$ of possible {\bf states} in which the system may be found in period $t$, for $t=1,...,n$;
\item the finite set $A_s$ of possible {\bf actions} that may be taken in state $s\in S_t$;
\item the {\bf transition probability} $p_{sj}^a$ from state $s\in S_t$ towards state $j\in S_{t+1}$, when action $a\in A_i$ is taken;
\item the {\bf expected immediate cost} $c_t(s,a)$ incurred if action $a\in A_s$ is taken in state $s\in S_t$ at the onset of period $t$;
\item the {\bf discount factor} $\alpha$;
\item the {\bf functional equation} $f_t(s)$ denoting the minimum {\bf expected total cost} incurred over periods $t,t+1,\ldots,n$, if the system is in state $s$ at the beginning of period $t$.
\end{itemize}
The decision maker's goal is to minimise the expected (discounted) total cost\footnote{or to maximise the expected (discounted)\index{discounting} total reward.} over the planning horizon. 
The functional equation typically takes the following structure
\[f_t(s)=\min_{a\in A_s} c_t(s,a) + \alpha \sum_{j\in S_{t+1}}p_{sj}^a f_{t+1}(j),\]
where the boundary condition of the system is $f_{n+1}(s)\triangleq 0$, for all $s\in S_{n+1}$. The goal is to determine $f_1(s)$, where $s$ is the state of the system at the beginning of the first period.

\subsection{A motivating example: stochastic inventory control}
Consider a 3-period inventory control problem. At the beginning of each period the firm should 
decide how many units of a product should be produced. If production takes place for $x$ units, where $x > 0$, 
we incur a production cost $c(x)= K+vx$, which comprises both fixed ($K=3$) and variable ($v=2$) components.
Production in each period cannot exceed 4 units. Demand in each period is 
1 or 2 units with equal probability (0.5). Demand is observed in each period only after production 
has occurred. After meeting current period's demand holding cost $h=1$ per unit is incurred 
for any item that is carried over from one period to the next. Because of limited capacity the inventory 
at the end of each period cannot exceed 3 units. All demand should be met on time (no backorders). 
If at the end of the planning horizon (i.e. period 3) the firm still has units in stock, these can 
be salvaged at cost $b=2$ per unit. The initial inventory is 1 unit.

We formulate the stochastic inventory control problem as a stochastic dynamic program:
\begin{itemize}
\item there are $n=4$ periods in the planning horizon;
\item the state $s\in S_t$ in period $t$ represents the initial inventory level at the beginning of period $t$, where $S_t=\{1,\ldots,3\}$, for $t=1,\ldots,n$;
\item the action $a$ given state $s$ in period $t$ is the order quantity $Q$, where $A_s=\{0,\ldots,4-s\}$;
\item the transition probability $p_{sj}^a$ from state $s\in S_t$ towards state $j\in S_{t+1}$, when action $a\in A_i$ is taken, immediately follows from the probability distribution of the demand $d$ in period $t$;
\item the expected immediate cost incurred if action $a\in A_s$ is taken in state $s\in S_t$ at the beginning of period $t=1,\ldots,3$ is 
\[c_t(s,a)=\left
\{
\begin{array}{ll}
K+v a+h\mbox{E}[\max(s+a-d,0)]	&a>0\\
h\mbox{E}[\max(s+a-d,0)]			&\mbox{otherw.}
\end{array}
\right.
\] 
finally, in period $t=4$, $c_4(s,Q)=b\mbox{E}[\max(s+a-d,0)]$, where $\mbox{E}$ denoted the expected value;
\item the discount factor $\alpha$ is 1;
\item the functional equation is $f_t(s)=\min c_t(s,a) + \mbox{E}[f_{t+1}(s+a-d)]$, with boundary condition $f_n(s)=\min c_4(s,a)$. The aim is to determine $f_1(1)$. 
\end{itemize}
Given this functional equation, an optimal policy can be obtained via forward recursion or backward recursion. In Section \ref{sec:sdp_java}, we shall present a novel approach for implementing a forward recursion algorithm using modelling constructs originally introduced in Java 8, which are next surveyed in Section \ref{sec:language_constructs}. This discussion will provide insights on the nature of the \texttt{jsdp} optimisation engine, which leverages the MapReduce framework originally discussed in \cite{10.1145/1327452.1327492} and seamlessly takes advantage of its parallelism and scalability.

\section{Language constructs}\label{sec:language_constructs}

In order to implement a compact forward recursion algorithm in Java we will rely on lambda expressions, functional interfaces, collections, and aggregate operations. We next survey each of these constructs in order.

\subsection{Lambda calculus and lambda expressions}

In computer programming, an anonymous function (also known as lambda function or lambda expression) is a function definition that is not bound to an identifier. Anonymous functions originate in the work of Church and in his invention of the lambda calculus \cite{Church_1940}. Java supports anonymous functions, named Lambda Expressions, starting with JDK 8. In Java, a lambda expression is a short block of code which takes in parameters and returns a value. Lambda expressions are similar to methods, but they do not need a name and they can be implemented right in the body of a method. The following lambda expression captures the immediate value function for a given state, action and demand value.
\begin{lstlisting}
(state, action, demand) -> {
            double cost = (action > 0 ? 3 + 2*action : 0);
            cost += h*(state.initialInventory+action-demand);
            cost -= (state.period == T ? b : 0)* (state.initialInventory+action-demand);
            return cost;
         };
\end{lstlisting}

\subsection{Functional interfaces}

In Java, any interface with a single abstract method is a functional interface, and its implementation may be treated as lambda expressions. The most simple and general case of a lambda is a functional interface with a method that receives one value and returns another. This function of a single argument is represented by the \texttt{Function} interface, which is parameterized by the types of its argument and a return value. For instance, we may use the following functional interface to capture the behaviour of a function that, given a state, generates an array of feasible actions, each of which is modelled as a \texttt{double} value.
\begin{lstlisting}
Function<State, double[]> actionGenerator;
\end{lstlisting}
More complex functional interfaces, which may receive or return more than one value, may be defined by leveraging the \texttt{@FunctionalInterface} annotation. For instance, we may define the following bespoke functional interface to capture a state transition function.
\begin{lstlisting}
@FunctionalInterface
   interface StateTransitionFunction <S, A, R> { 
      public S apply (S s, A a, R r);
   }
\end{lstlisting}

\subsection{Collections and aggregate operations}

The Java collections framework is a set of classes and interfaces that implement commonly reusable collection data structures. 

Collections are used to store objects and retrieve them efficiently. Rather than accessing elements in a collection using traditional \texttt{for} loop constructs, one may opt for aggregate operations instead. For instance, given a collection \texttt{actions}, one may print all actions in it as follows.
\begin{lstlisting}
actions.stream().forEach(a -> System.out.println(a));
\end{lstlisting}
A stream is a sequence of elements. Unlike a collection, it is not a data structure that stores elements. Instead, a stream carries values from a source through a pipeline. 
A pipeline is a sequence of aggregate operations, which contains the following components: a source collection, zero or more intermediate operations, which produce a new stream, and a terminal operation that produces a non-stream result, such as a primitive value (like a double value), a collection, or in the case of aggregate operation \texttt{forEach}, no value at all.

A key aggregate operation is the \texttt{map} function, which takes a lambda expression as argument, and returns a new stream consisting of the results of applying the given lambda expression to the elements of the stream. Java streams and aggregate operations find their origin in the seminal work \cite{10.1145/1327452.1327492} introducing the MapReduce framework. Aggregate operations \texttt{sum} and \texttt{average} are available for computing the respective functions. Finally, aggregate operator \texttt{filter} is used to return a stream that contains a subset of the elements in the original stream that meet a certain condition expressed as a lambda expression. 

\section{Stochastic Dynamic Programming in Java}\label{sec:sdp_java}

In this section, we illustrate how to implement a forward recursion algorithm by relying on modelling constructs introduced in Java 8 and illustrated in the previous section. The discussion in this section does not rely on the \texttt{jsdp} library; conversely, the aim here is to illustrate core modelling constructs and abstractions which \texttt{jsdp} is built upon.

To illustrate the concepts we are about to introduce, we will rely once more on the motivating stochastic inventory control problem previously introduced. To implement our forward recursion algorithm, we will rely on the following libraries.
\begin{lstlisting}
import java.util.Arrays;
import java.util.HashMap;
import java.util.Map;
import java.util.function.Function;
import java.util.stream.DoubleStream;
\end{lstlisting}

File \texttt{InventoryControl.java} embeds the following class, which will contain our code.
\begin{lstlisting}
public class InventoryControl {
  int planningHorizon;
  double[][] pmf;
  ...
}
\end{lstlisting}
Member variables \texttt{planningHorizon} denotes the number of periods in the planning horizon, while \texttt{pmf} is a two dimensional array that records a given probability mass function describing random demand in each period.

We define the following constructor.
\begin{lstlisting}
public class InventoryControl {
  ...
  public InventoryControl(int planningHorizon,
                          double[][] pmf) {
     this.planningHorizon = planningHorizon;
     this.pmf = pmf;
  }
  ...
}
\end{lstlisting}
Nested class \texttt{State} models the state of the system.
\begin{lstlisting}
public class InventoryControl {
  ...
  class State{
     int period, initialInventory;

     public State(int period, int initialInventory){
        this.period = period;
        this.initialInventory = initialInventory;
     }

     public double[] getFeasibleActions(){
        return actionGenerator.apply(this);
     }
  
     @Override
     public int hashCode(){
        String hash = "";
        hash = (hash + period) + "_" + this.initialInventory;
        return hash.hashCode();
     }

     @Override
     public boolean equals(Object o){
        if(o instanceof State)
           return  ((State) o).period == this.period &&
                   ((State) o).initialInventory == this.initialInventory;
        else
           return false;
     }

     @Override
     public String toString(){
        return this.period + " " + this.initialInventory;
     }
  }
  ...}
\end{lstlisting}
Method \texttt{hashCode()} is needed because we will store states in hashtables, which require each state to be uniquely identified by a hashcode for direct indexing; method \texttt{getFeasibleActions()} relies on actionGenerator, a function defined as follows.
\begin{lstlisting}
public class InventoryControl {
  ...
  Function<State, double[]> actionGenerator;
  ...}
\end{lstlisting}

One should recall that, for each state, we must be able to generate all feasible actions. For the moment, we leave \texttt{actionGenerator} unimplemented. We will later define an appropriate lambda expression that returns the appropriate set of actions for each relevant state.

In addition to the above functional interface we also define
\begin{lstlisting}
public class InventoryControl {
  ...
  @FunctionalInterface
  interface StateTransitionFunction <S, A, R> { 
    public S apply (S s, A a, R r);
  }

  public StateTransitionFunction<State, Double, Double> stateTransition;

  @FunctionalInterface
  interface ImmediateValueFunction <S, A, R, V> { 
     public V apply (S s, A a, R r);
  }

  public ImmediateValueFunction<State, Double, Double, Double> immediateValueFunction;
  ...}
\end{lstlisting}
capturing the state transition function, a function that, given a state, an action, and a random outcome, returns the associated future state; and the immediate value function, a function that, given a state, an action, and a random outcome, returns the associated immediate cost/profit.

We have now defined all relevant constructs that are necessary to set up our forward recursion procedure, which is presented in Figure \ref{fig:forward_recursion}.
\begin{figure*}
\begin{lstlisting}
public class InventoryControl {
  ...
  Map<State, Double> cacheActions = new HashMap<>();
  Map<State, Double> cacheValueFunction = new HashMap<>();
  double f(State state){
     return cacheValueFunction.computeIfAbsent(state, s -> {
        double val= Arrays.stream(s.getFeasibleActions())                            // Stream of feasible actions for state s
                                .map(orderQty -> Arrays.stream(pmf)                  // Stream of random variable support elements
                                          .mapToDouble(p -> p[1]*immediateValueFunction.apply(s, orderQty, p[0])+                           // Immediate cost
                                                                   (s.period < this.planningHorizon ?
                                                                   p[1]*f(stateTransition.apply(s, orderQty, p[0])) : 0)) // Recursion
                                          .sum())                                   // Compute probability weighted sum
                                .min()                                              // Compute optimal expected total cost
                                .getAsDouble(); 
        double bestOrderQty = Arrays.stream(s.getFeasibleActions())
                                             .filter(orderQty -> Arrays.stream(pmf)
                                                               .mapToDouble(p -> p[1]*immediateValueFunction.apply(s, orderQty, p[0])+
                                                                                       (s.period < this.planningHorizon ?
                                                                                        p[1]*f(stateTransition.apply(s, orderQty, p[0])):0))
                                                               .sum() == val)   // Find a cost-optimal action
                                    .findAny()
                                    .getAsDouble(); 
        cacheActions.putIfAbsent(s, bestOrderQty);                                 // Store optimal action in the cache
        return val;
     });
  }
  ...
}
\end{lstlisting}
\caption{Generic forward recursion procedure, which leverages lambda expressions, functional interfaces, collections, and aggregate operations}
\label{fig:forward_recursion}
\end{figure*}
the procedure directly implements the functional equation $f_t(s)$. It relies on memoization \cite{MICHIE_1968} --- method \texttt{computeIfAbsent()} --- to store the value of the functional equation for states that have been already visited. This ensures that states are not processed more than once.

Finally, we define our \texttt{main} method
\begin{lstlisting}
public class InventoryControl {
  ...
  public static void main(String [] args){
     int planningHorizon = 3;                //Planning horizon length
     double fixedProductionCost = 3;  //Fixed production cost
     double perUnitProductionCost = 2; //Per unit production cost
     int warehouseCapacity = 3;      //Production capacity
     double holdingCost = 1;             //Holding cost
     double salvageValue = 2;            //Salvage value
     int maxOrderQty = 4;                //Max order quantity
     double pmf[][] = {{1,0.5},{2,0.5}}; // Prob. mass fun.
     int maxDemand =  (int) Arrays.stream(pmf).mapToDouble(v -> v[0]).max().getAsDouble();
     int minDemand =  (int) Arrays.stream(pmf).mapToDouble(v -> v[0]).min().getAsDouble();
     ...
  }
  ...
}
\end{lstlisting}
and we introduce relevant lambda expressions that implement functional interfaces \texttt{actionGenerator}, \texttt{stateTransition}, and \texttt{immediateValueFunction}.
\begin{lstlisting}
public class InventoryControl {
  ...
  public static void main(String [] args){
    ...
    InventoryControl inventory = new InventoryControl(planningHorizon, pmf);
  
  
    /**
     * This function returns the set of actions associated 
     * with a given state
     */
    inventory.actionGenerator = state ->{
       int minQ = Math.max(maxDemand - state.initialInventory, 0);
       return DoubleStream.iterate(minQ, orderQty -> orderQty + 1)
                          .limit(Math.min(maxOrderQty, warehouseCapacity +  
                                          minDemand - state.initialInventory - minQ) + 1)
                          .toArray();
    };
  
    /**
     * State transition function; given a state, an action 
     * and a random outcome, the function returns the
     * future state
     */
    inventory.stateTransition = (state, action, randomOutcome) -> 
       inventory.new State(state.period + 1, (int) (state.initialInventory + action - randomOutcome));
  
    /**
     * Immediate value function for a given state
     */
    inventory.immediateValueFunction = (state, action, demand) -> {
        double cost = (action > 0 ? fixedProductionCost + perUnitProductionCost*action : 0);
        cost += holdingCost*(state.initialInventory+action-demand);
        cost -= (state.period == planningHorizon ? salvageValue : 0)*(state.initialInventory+action-demand);
        return cost;
    };
    ...      
  }...
}
\end{lstlisting}

We set up the problem parameters and we call relevant methods to obtain an optimal solution.
\begin{lstlisting}
public class InventoryControl {
  ...
  public static void main(String [] args){
     ...
    /**
     * Initial problem conditions
     */
     int initialPeriod = 1;
     int initialInventory = 1;
     State initialState = inventory.new State(initialPeriod, initialInventory);
  
    /**
     * Run forward recursion and determine the expected 
     * total cost of an optimal policy
     */
     System.out.println("f_1("+initialInventory+")=" + inventory.f(initialState));

    /**
     * Recover optimal action for period 1 when initial 
     * inventory at the beginning of period 1 is 1.
     */
     System.out.println("b_1("+initialInventory+")=" + inventory.cacheActions.get(inventory.new State(initialPeriod, initialInventory)));
  }...
}
\end{lstlisting}

After compiling and running the code the output obtained is \texttt{f\_1(1)=16.25}, and \texttt{b\_1(1)=3.0}.

\section{The \texttt{jsdp} library}\label{sec:jsdp}

The example presented in the previous section leveraged high level Java modelling constructs (lambda expressions, functional interfaces, collections and aggregate operators) to implement a forward recursion algorithm and tackle a specific stochastic dynamic programming problem in the realm of inventory control.

\texttt{jsdp} builds upon these modelling constructs and provides an additional layer of abstraction to model problems of decision making under uncertainty via Stochastic Dynamic Programming.

The library features off-the-shelf algorithms --- forward recursion and backward recursion --- as well as scenario reduction techniques to tackle generic problems of decision making under uncertainty with univariate or multivariate state descriptors.

In what follows, we will survey key modelling constructs offered by \texttt{jsdp} and we will demonstrate its flexibility on a well-known problem from stochastic inventory control \cite{Scarf1960}.

The key abstractions required to model a stochastic dynamic program are provided in package \texttt{jsdp.sdp}:
\begin{itemize}
\item \texttt{State}: this class is used to represent an abstract state;
\item \texttt{Action}: this class is used to represent an abstract action;
\item \texttt{TransitionProbability}: this class is used to represent transition probabilities;
\item \texttt{ImmediateValueFunction}: this functional interface is used to capture an abstract immediate value function;
\item \texttt{StateTransitionFunction} and \texttt{RandomOutcomeFunction}: these functional interfaces are used to capture an abstract state transition function.
\end{itemize}

The abstraction for the functional equation is provided by class \texttt{Recursion}, which generalises classes \texttt{BackwardRecursion} and \texttt{ForwardRecursion}, which in turn implement general purpose backward and forward recursion algorithms, respectively.

In order to model a problem with \texttt{jsdp} one may define concrete implementations of abstract classes in package \texttt{jsdp.sdp}. An example of this approach is given in package \texttt{jsdp.app.lotsizing}. This is a cumbersome solution that should be adopted only for complex problems.

In most cases, it is sufficient to rely upon general purpose concrete implementations provided in package \texttt{jsdp.sdp.impl}. In what follows, we shall illustrate this latter solution. A generic skeleton for a stochastic dynamic program developed in \texttt{jsdp} is provided in package \texttt{jsdp.app.skeleton}, the following example extends this skeleton.

\subsection{Scarf's stochastic lot sizing problem}\label{sec:scarf}

In this section, we provide a \texttt{jsdp} implementation for modelling and solving the well-known stochastic lot sizing problem investigated by Scarf in \cite{Scarf1960}. 

We consider an $n$-period inventory control problem. At the beginning of each period the firm should decide how many units of a product should be produced. If production takes place for $x$ units, where $x > 0$, we incur a production cost $c(x)$. This cost comprises both a fix $K$ and a variable $v$ component: 
\[c(x)=\left
\{
\begin{array}{ll}
K+vx		&x>0\\
0		&\mbox{otherwise}.
\end{array}
\right.
\] 
The order is delivered immediately at the beginning of the period. Demand in each period $t$ is Poisson distributed with known mean. Demand is observed in each period only after production has occurred. After meeting current period's demand holding cost $h$ per unit is incurred for any item that is carried over from one period to the next. Unmet demand in any given period is backordered at cost $b$ per unit per period. The initial inventory is known.


We next illustrate how to model and solve this stochastic inventory control problem in \texttt{jsdp}.

In file \texttt{InventoryControl.java}, we create a class \texttt{InventoryControl} and build a \texttt{main} method as follows.

\begin{lstlisting}
public class InventoryControl {
  ...
  public static void main(String [] args){
     ...
  }  
  ...
}
\end{lstlisting}
\newpage
First, we define problem parameters.
\begin{lstlisting}
/****************************************************
 * Problem parameters
 */
 
double fixedOrderingCost = 300; 
double proportionalOrderingCost = 0; 
double holdingCost = 1;
double penaltyCost = 10;
  
double[] meanDemand = {10,20,15,20,15,10};
double coefficientOfVariation = 0.2;
\end{lstlisting}

We next define the quantile at which probability distributions used will be truncated.
\begin{lstlisting}
double truncationQuantile = 0.999;
\end{lstlisting}

We introduce probability distributions of the random variables in our model.
\begin{lstlisting}
// Random variables

Distribution[] distributions = IntStream.iterate(0, i->i+1)
                                        .limit(meanDemand.length)
                                        .mapToObj(i -> new PoissonDist(meanDemand[i]))
                                        .toArray(Distribution[]::new);
\end{lstlisting}

We create two arrays storing the lower bound and the upper bound of random variable supports.
\begin{lstlisting}
double[] supportLB = IntStream.iterate(0, i -> i + 1)
                              .limit(meanDemand.length)
                              .mapToDouble(i -> distributions[i].inverseF(1 - truncationQuantile))
                              .toArray();
double[] supportUB = IntStream.iterate(0, i -> i + 1)
                              .limit(meanDemand.length)
                              .mapToDouble(i -> distributions[i].inverseF( truncationQuantile))
                              .toArray();
\end{lstlisting}

We define a variable representing the initial inventory level.
\begin{lstlisting}
double initialInventory = 0;
\end{lstlisting}

We then proceed to the model definition; the first step is to characterize the state space.
\begin{lstlisting}
/*****************************************************
 * Model definition
 */

// State space
  
double stepSize = 1;       //Stepsize must be 1 for discrete distributions
double minState = -50;     //Inventory level lower bound in each period
double maxState = 150;     //Inventory level upper bound in each period
\end{lstlisting}

The following instruction bounds the state space.
\begin{lstlisting}
StateImpl.setStateBoundaries(stepSize, minState, maxState);
\end{lstlisting}

We introduce a functional interface that dynamically computes the \texttt{ArrayList<Action> feasibleActions} storing feasible actions for a given state \texttt{s}.
\begin{lstlisting}
// Actions
  
Function<State, ArrayList<Action>> buildActionList = s -> {
   StateImpl state = (StateImpl) s;
   ArrayList<Action> feasibleActions = new ArrayList<Action>();
   for(double i = state.getInitialState(); 
       i <= StateImpl.getMaxState(); 
       i += StateImpl.getStepSize()){
      feasibleActions.add(new ActionImpl(state, i - state.getInitialState()));
   }
   return feasibleActions;
};
\end{lstlisting}

The next functional interface defines the idempotent action: an action that does not affect the state variable.
\begin{lstlisting}
Function<State, Action> idempotentAction = s -> new ActionImpl(s, 0.0);
\end{lstlisting}

The following two functional interfaces define the immediate value function
\begin{lstlisting}
// Immediate Value Function
  
ImmediateValueFunction<State, Action, Double> immediateValueFunction = (initialState, action, finalState) -> {
   ActionImpl a = (ActionImpl)action;
   StateImpl fs = (StateImpl)finalState;
   double orderingCost = 
         a.getAction() > 0 ? (fixedOrderingCost + a.getAction()*proportionalOrderingCost) : 0;
   double holdingAndPenaltyCost =   
         holdingCost*Math.max(fs.getInitialState(),0) + penaltyCost*Math.max(-fs.getInitialState(),0);
   return orderingCost+holdingAndPenaltyCost;
};
\end{lstlisting}
and the random outcome function, 
\begin{lstlisting}
// Random Outcome Function
  
RandomOutcomeFunction<State, Action, Double> randomOutcomeFunction = (initialState, action, finalState) -> {
   double realizedDemand = ((StateImpl)initialState).getInitialState() +
                           ((ActionImpl)action).getAction() -
                           ((StateImpl)finalState).getInitialState();
   return realizedDemand;
};
\end{lstlisting}
which --- given present state, future state, and action chosen --- returns the associated random outcome. 
This is essentially equivalent to defining the dynamics --- i.e. state transition function --- of the system; adopting a random outcome function in lieu of a state transition function is useful if one aims to rely on sampling for speeding up the computation and obtain an approximate policy.

To summarize, in order to model a problem the key steps are the following: bound the state space, define the functional interfaces to compute the action list, the immediate value function, and the random outcome function for the system.

Once these modeling components are defined, we can move to the solution process.
The first step is to determine if we want to sample the state space or not.
\texttt{SamplingScheme.NONE} carries out an exhaustive state space enumeration; 
in the following example we will adopt simple random sampling.
\begin{lstlisting}
/**************************************************
 * Solve
 */
  
SamplingScheme samplingScheme = SamplingScheme.SIMPLE_RANDOM_SAMPLING; 
int maxSampleSize = 100;
double reductionFactorPerStage = 1;
\end{lstlisting}
The maximum sample size (\texttt{maxSampleSize}) refers to the sample size adopted for random variables at stage 1; if the reduction factor per stage (\texttt{reductionFactorPerStage}) is 1, this sample size will be adopted also in subsequent stages, otherwise the sample size in subsequent stages will shrink exponentially fast according to the rule
\begin{lstlisting}
Math.ceil(maxSamples/Math.pow(reductionFactorPerStage, period));
\end{lstlisting}
where \texttt{period} denotes the stage for which we are computing the the sample size. We name this approach {\bf sample waning}. To the best of our knowledge, a similar approach has not yet been discussed in the literature. It should be noted that the appeal of sample waning in the context of a forward or backward recursion algorithm is related to the advantage brought by memoization. 

Finally, we proceed and apply backward recursion to solve our problem. The discount factor (\texttt{discountFactor}) captures value discounting from one period to the next. \texttt{stateSpaceLowerBound} and \texttt{loadFactor} are parameters utilised to initialise \texttt{hashtables}. In this specific instance, we use \texttt{HashType.THASHMAP} as hash table to store the state space. Enum \texttt{HashType} contains other possible choices, including a hash table (\url{https://mapdb.org/}) that provide disk storage in place of RAM for large state spaces.
\begin{lstlisting}
// Value Function Processing Method: backward recursion

double discountFactor = 1.0;
int stateSpaceLowerBound = 10000000;
float loadFactor = 0.8F;
BackwardRecursionImpl recursion = new BackwardRecursionImpl(OptimisationDirection.MIN,
                                    distributions,
                                    supportLB,
                                    supportUB,
                                    immediateValueFunction,
                                    randomOutcomeFunction,
                                    buildActionList,
                                    idempotentAction,
                                    discountFactor,
                                    samplingScheme,
                                    maxSampleSize,
                                    reductionFactorPerStage,
                                    stateSpaceLowerBound,
                                    loadFactor,
                                    HashType.THASHMAP);
\end{lstlisting}

Backward recursion is invoked as follows.
\begin{lstlisting}
System.out.println("---Backward recursion---");

recursion.runBackwardRecursionMonitoring();

double ETC = recursion.getExpectedCost(initialInventory);
StateDescriptorImpl initialState = new StateDescriptorImpl(0, initialInventory);
double action = recursion.getOptimalAction(initialState).getAction();
long percent = recursion.getMonitoringInterfaceBackward().getPercentCPU();

System.out.println("Expected total cost: "+ETC);
System.out.println("Optimal initial action: "+action);
System.out.println("Time elapsed: "+recursion.getMonitoringInterfaceBackward().getTime());
System.out.println("CPU usage: "+percent+"% ("+Runtime.getRuntime().availableProcessors()+" cores)");
\end{lstlisting}

After running the code the output obtained is the following.
\begin{lstlisting}
---Backward recursion---
Expected total cost: 567.7537178866613
Optimal initial action: 91.0
Time elapsed: 10
CPU usage: 168% (4 cores)
\end{lstlisting}
\texttt{jsdp} exploits parallelisation made available off-the-shelf by Java streams, this is evidenced in the CPU usage statistics. 

\section{Applications}\label{sec:applications}


\texttt{jsdp} has been used in a number of published works to generate optimal solutions to large multi-stage problems against which heuristics have been benchmarked, see e.g. \cite[Section 4.1]{DuralSelcuk2020}. In the rest of this section, we shall focus on a stochastic vehicle routing problem (package \texttt{jsdp.app.routing}), which was originally discussed in \cite{Rossi_2019}. We will present new results to illustrate the effectiveness of the sample waning heuristic discussed in the previous section.

We consider the Stochastic Bowser Routing Problem discussed in \cite{Rossi_2019}, and in particular, the test bed discussed in Section 5.2.1, which comprises a total of 108 instances. 
We set the initial sample size to 30 and the waning factor to 5. 

We implement the solution obtained using sample waning by following a Receding Horizon (RH) approach \cite{Bellingham2002}. 
All results presented are averages obtained over 500 replications of RH. To account for estimation errors, our results reflect the upper limit of the associated 95\% confidence interval for the mean obtained; therefore they are conservative figures. 

The average cost difference between the heuristic and the exact stochastic models, in percentage increase over the cost of the optimal stochastic solution, is 23.4\%, the median cost increase is 7.65\%, the maximum cost increase is 176\%. A histogram detailing the percentage of instances featuring a given optimality gap is shown in Fig. \ref{fig:sdp_fuel_optimality_gap_histograms}. The average state space size decreased from 475517 to 9785; the average solution time decreased from 48 minutes to 50 seconds. 

\captionsetup[figure]{font=scriptsize,labelfont=scriptsize}

\begin{minipage}{\linewidth}
\begin{minipage}{.42\textwidth}
\begin{figure}[H]
\resizebox{\textwidth}{!}{
\begin{tikzpicture}
\begin{axis}[
    xlabel=Optimality gap (\%),
    ylabel=Percentage of instances,
    ybar,
    ymin=0
]
\addplot +[
    hist={
        bins=30,
        data min=0,
        data max=180
    }   
] table [y index=0] {data_sdp_fuel.csv};
\end{axis}
\end{tikzpicture}
}
\caption{Stochastic fuel consumption}
\label{fig:sdp_fuel_optimality_gap_histograms}
\end{figure}
\end{minipage}
\hspace{0.05\linewidth}
\begin{minipage}{.42\textwidth}
\begin{figure}[H]
\resizebox{\textwidth}{!}{
\begin{tikzpicture}
\begin{axis}[
    xlabel=Optimality gap (\%),
    ylabel=Percentage of instances,
    ybar,
    ymin=0
]
\addplot +[
    hist={
        bins=30,
        data min=0,
        data max=30
    }   
] table [y index=0] {data_sdp_location.csv};
\end{axis}
\end{tikzpicture}
}
\caption{Stochastic asset location}
\label{fig:sdp_location_optimality_gap_histograms}
\end{figure}
\end{minipage}
\end{minipage}

\vspace{0.5em}

\captionsetup[figure]{font=small,labelfont=small}

We carried out a similar set of experiments under uncertain asset location; asset fuel consumption is now deterministic and equal to the mean consumption value in each period, while asset location probabilities have been generated by randomly selecting for each period two possible site locations and by assigning a probability of 0.5 to each of them --- note that the two locations can incidentally be the same.

The average cost difference between the heuristic and the exact stochastic models, in percentage increase over the cost of the optimal stochastic solution, is 2.90\%, the median cost increase is 1.80\%, the maximum cost increase is 23.3\%. A histogram detailing the percentage of instances featuring a given optimality gap is shown in Fig. \ref{fig:sdp_location_optimality_gap_histograms}. The average state space size decreased from 7245 to 1307; the average solution time decreased from 2.25 to 2.02 seconds. 

These experiments expose the tradeoff between higher computational savings --- as in the original stochastic fuel consumption case --- leading to lower quality solutions; and more modest computational savings --- as in the stochastic asset location case --- leading to higher quality solutions. Computational savings and solution quality can be controlled by varying the initial sample size and the waning factor.

\section{Related works and conclusions}\label{sec:related_works}

There exist several commercial algebraic modelling languages (AML) available in the market (e.g. AMPL, OPL). These languages, also known as ``Mathematical Programming'' (MP) languages, let a user express a mathematical model by using conventional mathematical notation. A compiler translates these high level models to a form understandable by MP solvers (such as CPLEX), which then find an optimal solution. The counterpart of MP for modelling problems of decision making under uncertainty is Stochastic Programming (SP). In addition to problem parameters, decision variables, constraints and an objective function, SP features constructs for modelling uncertainty: e.g. random variables, stochastic constraints, and decision stages. Over fifteen years ago, \cite{1105.90337} discussed the fact that most of the difficulties to model uncertainty through SP originate from the lack of an agreed standard of representation and of a widely accepted syntax for a description of stochastic programs. This message has recently been reinforced in \cite{Powell_2016}. 
In the past decade significant research and commercial efforts have been directed towards the development of effective modelling languages for stochastic programming, i.e. Stochastic AML (SAML).
Stochastic extensions exist for several well-known AMLs: SMPS \cite{citeulike:10985899}, SAMLP \cite{Fourer:2009:SFM:1538318.1538324}, XPress-SP \cite{Geunes05}, Stochastic GAMS and AIMMS, Pyomo AML \cite{pysp,osti_1721736}, Stochastic OPL \cite{citeulike:584318}, and StochasticPrograms.jl \cite{Biel2022}. 
All these extensions adopt a scenario-based modelling paradigm, which is not suitable for multi-stage problems featuring long planning horizons, since the nodes in the scenario tree grow exponentially with the number of stages. Conversely, the state space of a stochastic dynamic program grows linearly in the number of stages.
Libraries such as ApproxRL and OpenAI have been proposed for Reinforcement Learning \cite{Sutton1998}, in which however the environment is modelled as a Markov Decision Process; \texttt{jsdp} does not require the decision process to be Markovian.
There are limited options when it comes to high level declarative modelling and solution frameworks for problems of decision making under uncertainty formulated as stochastic dynamic programs. 
Libraries such as SDDP.jl \cite{Dowson2021} have been recently proposed for solving multi-stage stochastic programming problems using Stochastic Dual Dynamic Programming, which recursively constructs an approximation of the problem functional equation through linear programming duality, and requires a convex cost structure and linear dynamics. \texttt{jsdp} is complete, leaner, and more general: it directly computes the functional equation without relying on a linear programming solver, and hence it does not operate under such restrictions.

\bibliographystyle{IEEEtran}
\bibliography{IEEEabrv,bibliography}

\end{document}